# The Factored Frontier Algorithm for Approximate Inference in DBNs

**Kevin Murphy and Yair Weiss**
Computer Science Department
University of California
Berkeley, CA 94720-1776
{murphyk,yweiss}@cs.berkeley.edu

## Abstract

The Factored Frontier (FF) algorithm is a simple approximate inference algorithm for Dynamic Bayesian Networks (DBNs). It is very similar to the fully factorized version of the Boyen-Koller (BK) algorithm, but instead of doing an exact update at every step followed by marginalisation (projection), it always works with factored distributions. Hence it can be applied to models for which the exact update step is intractable. We show that FF is equivalent to (one iteration of) loopy belief propagation (LBP) on the original DBN, and that BK is equivalent (to one iteration of) LBP on a DBN where we cluster some of the nodes. We then show empirically that by iterating more than once, LBP can improve on the accuracy of both FF and BK. We compare these algorithms on two real-world DBNs: the first is a model of a water treatment plant, and the second is a coupled HMM, used to model freeway traffic.

## 1 Introduction

Dynamic Bayesian Networks (DBNs) are directed graphical models of stochastic processes. They generalise hidden Markov models (HMMs) by representing the hidden (and observed) state in terms of state variables, which can have complex interdependencies. The graphical structure provides an easy way to specify these conditional independencies, and hence to provide a compact parameterization of the model. See Figure 1 for some examples.

In this paper, we will be concerned with the task of offline probabilistic inference in DBNs, i.e., computing $P(X_t^i|y_{1:T})$ for $t = 1, \ldots, T$ and $i = 1, \ldots, N$, where $X_t^i$ is the $i$'th hidden node at time $t$, and $y_t$ is evidence vector at time $t$; this is often called "smoothing". We will assume that all the hidden nodes are discrete and each has $Q$ possible values. The observed nodes can be discrete or continuous.

The simplest way to perform exact inference in a DBN is to convert the model to an HMM and apply the forwards-backwards algorithm [18]. This takes $O(TQ^{2N})$ time. By exploiting the conditional independencies within a slice, it is possible to reduce this to $\Omega(TNQ^{N+F})$ time, where $F$ is the maximum fan-in of any node. Unfortunately, this is still exponential in $N$. In fact, this is nearly always the case (assuming the graph is connected), because even if there is no direct connection between two nodes in the same or neighboring "time slices," they will become correlated over time by virtue of sharing common influences in the past. Hence, unlike the case for static networks, we need to use approximations even for "sparse" models.

In Section 3.1, we present a new approximation, called the "factored frontier" (FF) algorithm, which represents the belief state as a product of marginals. The FF algorithm is thus very similar to the fully factorized version of the Boyen-Koller (BK) algorithm, which we summarise in Section 3.2. FF, however, is a more aggressive approximation, and can therefore be applied when even BK is intractable: FF will always take $O(TNQ^{F+1})$ time, whereas BK can take more, depending on the graph.

In Section 4, we show how both FF and BK are related to loopy belief propagation (LBP) [15, 21, 20, 6, 7, 13], which is the method of applying Pearl's message passing algorithm [16] to a Bayes net even if it contains (undirected) cycles or loops. In Section 5, we experimentally compare all four algorithms — exact, FF, BK, and LBP — on a number of problems, and in Section 7, we conclude.

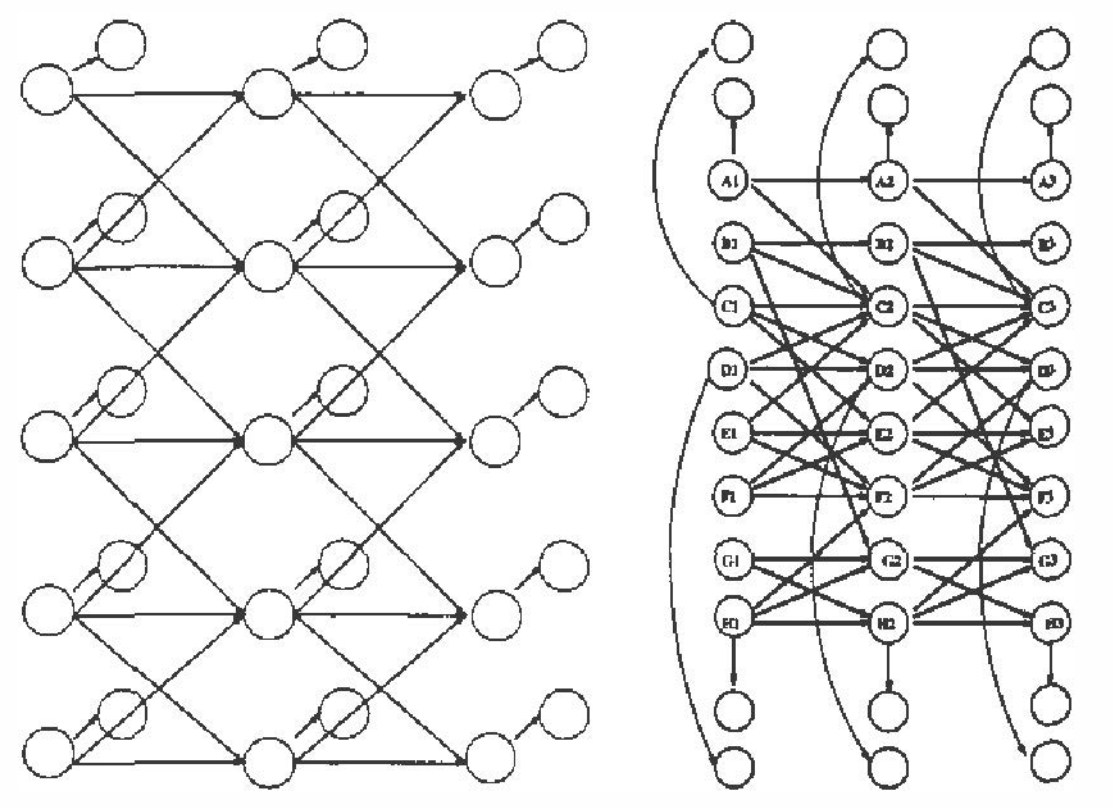

a b

Figure 1: Some DBNs. (a) A coupled HMM with $N = 5$ chains and $T = 3$ timeslices. Clear nodes are hidden, shaded nodes are observed. In the freeway traffic application in Section 5, $X_t^i$ represents the hidden traffic status (free-flowing or congested) at location $i$ on the freeway at time $t$; this is assumed to generate a local noisy measurement of traffic speed, $Y_t^i$, and to depend on its previous state and the previous state of its upstream and downstream neighbors. (b) A DBN designed to monitor a waste water treatement plant. This model is originally from [9], and was modified by [2] to include (discrete) evidence nodes.

## 2 Exact inference

We start by reviewing the forwards-backwards (FB) algorithm [18] for HMMs, and then the frontier algorithm [23] for DBNs, since this will form the basis of our generalisation.

### 2.1 The forwards backwards algorithm

The basic idea of the FB algorithm is to compute $\alpha_t^i \stackrel{\text{def}}{=} P(X_t = i|y_{1:t})$ in the forwards pass, $\beta_t^i \stackrel{\text{def}}{=} P(y_{t+1:T}|X_t = i)$ in the backwards pass, and then to combine them to produce the final answer: $\gamma_t^i \stackrel{\text{def}}{=} P(X_t = i|y_{1:T}) \propto \alpha_t^i \beta_t^i$. Let $M(i,j) \stackrel{\text{def}}{=} P(X_{t+1} = j|X_t = i)$ be the transition matrix, and $W_t(i,i) \stackrel{\text{def}}{=} P(y_t|X_t = i)$ be a diagonal matrix containing the conditional likelihood of the evidence at time $t$. The algorithm is just repated matrix-vector multiplication. Specifically, in the forwards pass we compute $\alpha_t \propto W_t M^T \alpha_{t-1}$, and in the backwards pass we compute $\beta_t \propto M W_{t+1} \beta_{t+1}$; the constants of proportionality are simply the normalizing constants. The boundary conditions are $\alpha_1 = W_1 \pi$ and $\beta_T = 1$, where $\pi^i \stackrel{\text{def}}{=} P(X_1 = i)$ is the prior. If $X$ can be in $S$ possible states, the FB algorithm clearly takes $O(S^2 T)$ time.

### 2.2 The frontier algorithm

If $X_t$ is a vector of $N$ hidden nodes, each with $Q$ possible values, then $X$ can be in $S = Q^N$ possible states, so the FB algorithm becomes intractable. The frontier algorithm [23] is a way of computing $\alpha_t$ from $\alpha_{t-1}$ (and similarly for the $\beta_t$'s) without needing to form the $Q^N \times Q^N$ transition matrix, yet alone multiply by it.

The basic idea is to "sweep" a Markov blanket across the DBN, first forwards and then backwards. We shall call the nodes in the Markov blanket the "frontier set", and denote it by $\mathcal{F}$; the nodes to the left and right of the frontier will be denoted by $\mathcal{L}$ and $\mathcal{R}$. At every step of the algorithm, $\mathcal{F}$ d-separates $\mathcal{L}$ and $\mathcal{R}$. We will maintain a joint distribution over the nodes in $\mathcal{F}$.

We can advance the frontier from slice $t-1$ to $t$ as follows. We move a node from $\mathcal{R}$ to $\mathcal{F}$ as soon as all its parents are in $\mathcal{F}$. To keep the frontier as small as possible, we move a node from $\mathcal{F}$ to $\mathcal{L}$ as soon as all its children are in $\mathcal{F}$. Adding a node entails multiplying its conditional probability table (CPT) $P(X_t^i|\text{Pa}(X_t^i))$ onto the frontier, and removing a node entails marginalising it out of the frontier.

This is best explained by example (see Figure 2). Consider the coupled HMM (CHMM) shown in Figure 1. The frontier initially contains all the nodes in slice $t-1$: $F_{t,0} \stackrel{\text{def}}{=} \alpha_{t-1} = P(X_{t-1}^{1:N}|y_{1:t-1})$. We then advance the frontier by moving $X_t^1$ from $\mathcal{R}$ to $\mathcal{F}$. To do this, we multiply in its CPT $P(X_t^1|X_{t-1}^1, X_{t-1}^2)$:

$$F_{t,1} = P(X_t^1, X_{t-1}^{1:N}|y_{1:t-1}) = P(X_t^1|X_{t-1}^1, X_{t-1}^2) \times F_{t,0}$$

Next we add in $X_t^2$:

$$\begin{aligned} F_{t,2} &= P(X_t^{1:2}, X_{t-1}^{1:N}|y_{1:t-1}) \\ &= P(X_t^2|X_{t-1}^1, X_{t-1}^2, X_{t-1}^3) \times F_{t,1} \end{aligned}$$

Now all of the nodes that depend on $X_{t-1}^1$ are in the frontier, so we can marginalize $X_{t-1}^1$ out (move it from $\mathcal{F}$ to $\mathcal{L}$):

$$F_{t,3} = P(X_t^{1:2}, X_{t-1}^{2:N}|y_{1:t-1}) = \sum_{X_{t-1}^1} F_{t,2}$$

The process continues in this way until we compute

$$F_{t,N} = P(X_t^{1:N}|y_{1:t-1})$$

Finally, we weight this factor by the likelihood:

$$\alpha_t = P(X_t^{1:N}|y_{1:t}) \propto P(y_t|X_t^{1:N}) \times F_{t,N}$$

It is clear that in this example, exact inference takes $O(TNQ^{N+2})$ time and space, since the frontier never contains more than $N + 2$ nodes, and it takes $O(N)$ steps to sweep the frontier from $t-1$ to $t$. In general, the running time of the frontier algorithm is exponential in the size of the largest frontier; this quantity is also known as the induced width of the underlying or moral graph. We would therefore like to keep the frontiers as small as possible. Unfortunately, computing an order in which to add and remove nodes so as to minimize the sum of the frontier sizes is equivalent to finding an optimal elimination ordering, which is known to be NP-hard. Nevertheless, heuristics methods, such as greedy search [10], often perform as well as exhaustive search using branch and bound [23].

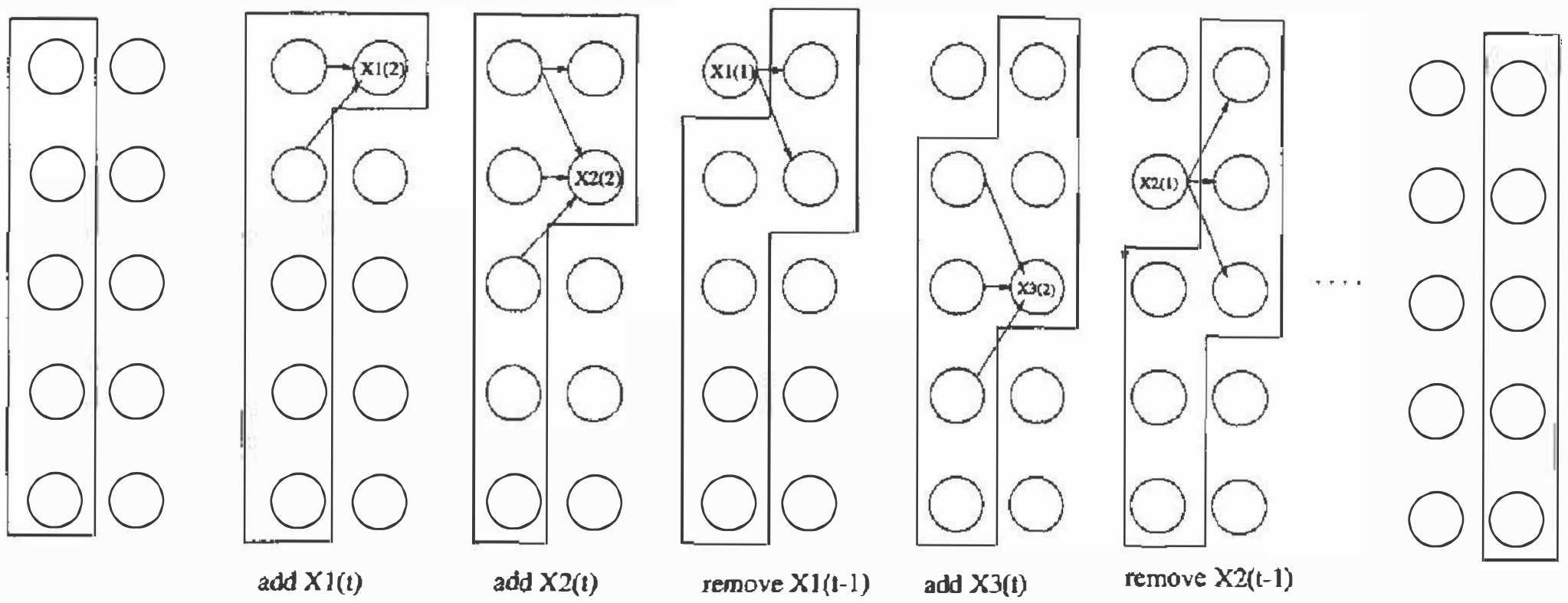

Figure 2: The frontier algorithm applied to a CHMM; observed leaves are omitted for clarity. Nodes inside the box are in the frontier. The node being operated on is shown shaded; only connections with its parents and children are shown; other arcs are omitted for clarity. See text for details.

A special case of the frontier algorithm, applied to factorial HMMs, was published in Appendix B of [8]. (In an FHMM, there are no cross links between the hidden nodes, so there are no constraints on the order in which nodes are added to or removed from the frontier.) For regular[1] DBNs, the frontier algorithm is equivalent to the junction tree algorithm [3, 11, 19] applied to the "unrolled" DBN. In particular, the frontier sets correspond to the maximal cliques in the moralized, triangulated graph; in the junction tree, these cliques are connected together in a chain, possibly with some smaller cliques "hanging off the backbone" to accomodate the non-persistent observed leaves. Despite this equivalence to junction tree, the frontier algorithm is appealingly simple, and will form the basis of the approximation algorithm discussed in the next section.

[1]A regular DBN has certain restrictions on its topology. Let $H_t$ denote all the hidden nodes in time-slice $t$, and $O_t$ all the observed nodes. A regular DBN can have connections from $H_t$ to $O_t$ and to $H_{t+1}$, but to nowhere else. In particular, there cannot be any intra-slice connections within the $H_t$ nodes. Furthermore, we assume each node in $H_t$ connects to one or more nodes in $H_{t+1}$ (i.e., is persistent). All the DBNs in this paper are regular.

The frontier algorithm works for non-regular DBNs, but it may be less efficient that junction tree in this case. The factored frontier and loopy belief propagation algorithms also work for non-regular DBNs.

# 3 Approximate inference

## 3.1 The factored frontier algorithm

The problem with the frontier and junction tree algorithms is that they need exponential space just to *represent* the belief states, and hence need at least that much time to compute them. The idea of the factored frontier (FF) algorithm is to approximate the belief state with a product of marginals: $P(X_t|y_{1:t}) \approx \prod_{i=1}^{N} P(X_t^i|y_{1:t})$. (The backward messages $\beta_t$ are approximated in a similar way.)

The algorithm proceeds as follows: when we add a node to the frontier, we multiply its CPT by the product of the factors corresponding to its parents; this creates a joint distribution for this family. We then immediately marginalize out the parent nodes. The backwards pass is analogous. This is like the frontier algorithm except that we always maintain the joint distribution over the frontier nodes in factored form. This algorithm clearly takes $O(TNQ^{F+1})$ time, no matter what the topology.

## 3.2 The Boyen-Koller algorithm

The Boyen-Koller algorithm [2] represents the belief state, $\alpha_t = P(X_t|y_{1:t})$, as a product of marginals over $C$ "clusters", $P(X_t|y_{1:t}) \approx \prod_{c=1}^{C} P(X_t^c|y_{1:t})$, where $X_t^c$ is a subset of the variables $\{X_t^i\}$. (The clusters do not need to be disjoint.) Given a factored prior, $\tilde{\alpha}_{t-1}$, we do one step of exact Bayesian updating to compute the posterior, $\hat{\alpha}_t$. In general, $\hat{\alpha}_t$ will not be factored as above, so we need to project to the space of factored distributions by computing the marginal on each clus-ter. The product of these marginals then gives the approximate posterior, $\tilde{\alpha}_t$. We can use a similar method for computing the backward messages in an efficient manner [1]. Boyen and Koller prove, roughly speaking, that if the error introduced by the projection step isn't much greater than the error incurred by using an approximate prior, both errors relative to the true (uncomputable) distribution, then the overall error is bounded.

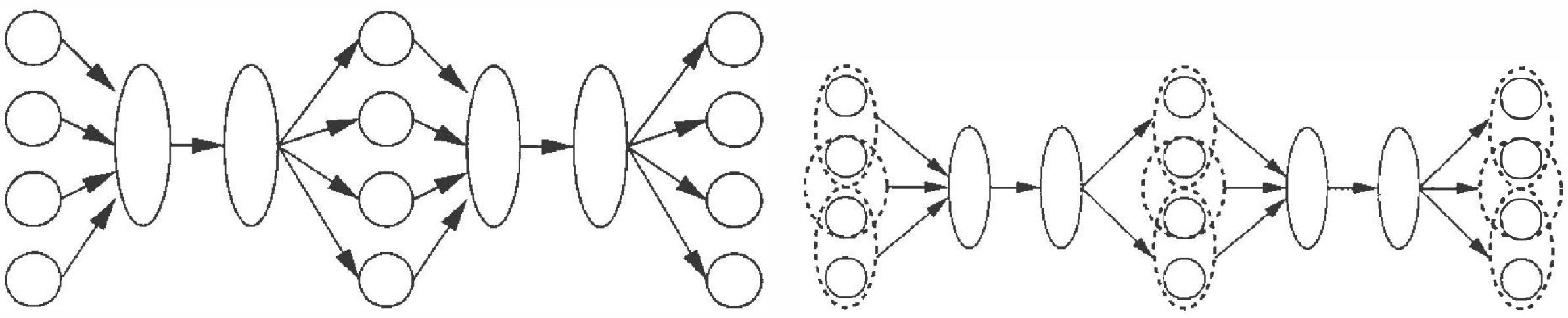

Figure 3: Illustration of the clustering process. (a) This is a modified version of a CHMM with 4 chains. The big "mega nodes" contain the joint distribution on the whole slice. We have omitted the observed leaves for clarity. LBP applied to this graph is equivalent to BK. (b) This is like (a), except we have created overlapping clusters of size 2, for additional accuracy.

The accuracy of the BK algorithm depends on the size of the clusters that we use to approximate the belief state. Exact inference corresponds to using a single cluster, containing all the hidden variables in a time-slice. The most aggressive approximation corresponds to using $N$ clusters, one per variable; we call this the "fully factorized" approximation.

It is clear that the fully factorized version of BK is very similar to the FF algorithm, but there is one important difference: BK assumes that we update the factored prior exactly (using, say, junction tree) before computing the marginals, whereas FF computes the (approximate) marginals directly. BK is obviously more accurate than FF, but sometimes it cannot be used, because even one step of exact updating is too expensive.

The cost of using BK is determined by the size of the maximal cliques of the moralized, triangulated version of the *two-slice* DBN. (Unrolling the DBN for many slices induces long-distance correlations, and results in cliques that span the whole time-slice, as we saw above.) For the coupled HMM (CHMM) model in Figure 1, the cliques just correspond to the families (nodes and their parents), so the algorithm takes $O(TNQ^{F+1})$ time, the same as FF. But for the water model (see Figure 1), we also get extra "non-local" cliques due to triangulation. For more complex models, such as the 2D generalisation of a CHMM — where each time slice is now an $N = n \times n$ lattice, and each cell depends on all the nodes in its "receptive field" in the previous slice (but not on its neighbors within a slice) — the largest clique has size $n$, and hence the running time of BK is $O(TNQ^{\sqrt{N}})$, even in the fully factorized case.

## 4 BK and FF as special cases of loopy belief propagation

Pearl's belief propagation algorithm [16] is a way of computing exact marginal posterior probabilities in graphs with no undirected cycles (loops). Essentially it generalises the forwards-backwards algorithm to trees. When applied to a graph with loops, the algorithm is sometimes called "loopy belief propagation" (LBP); in this case, the resulting "posteriors" may not be correct, and can even oscillate. Nevertheless, the outstanding empirical success of turbo decoding, which has be shown to be equivalent to LBP [13], has created great interest in the algorithm.

LBP has been empirically shown to work well on several kinds of Bayesian networks which are quite different from turbo codes [15, 7]. In addition, a number of theoretical results have now been proved for networks in which all nodes are Gaussian [21], for networks in which there is only a single loop [20], and for general networks but using the max-product (Viterbi) version instead of the sum-product (forwards-backwards) version of the algorithm [6].

The key assumption in LBP is that the messages coming into a node are independent. But this is exactly the same assumption that we make in the FF algorithm! Indeed, we can show that both algorithms are equivalent if we use a specific order in which to send messages. Normally we implement LBP using a decentralized message passing protocol, in which, at each step, every node computes its own $\lambda$ and $\pi$ in parallel (based on the incoming message at the previous step), and then sends out $\lambda$ and $\pi$ messages to all its neighbors. However, we can also imagine a forwards-backwards (FB) protocol, in which each node first sends $\pi$ ($\alpha$) messages from left to right, and then sends $\lambda$ ($\beta$) messages from right to left. A single pass of this FB protocol is equivalent to FF.[2]

The fixed points of LBP are the same, no matter what protocol is used. If there is not a unique fixed point, the algorithms may end up at different answers. They can also have different behavior in the short term. In particular, if the DBN is in fact an HMM, then a single FB iteration ($2TN$ message computations) will result in the exact posteriors, whereas it requires $T$ iterations of the decentralized protocol (each iteration computing $2TN$ messages in parallel) to reach the same result; hence the centralized algorithm is more efficient [17]. For loopy graphs, it is not clear which protocol is better; it depends on whether local or global information is more important for computing the posteriors. In this paper, we use the centralized (FB) protocol.

It is also easy to see that the fully-factorized version of BK is equivalent to a single FB pass of LBP applied to a modified DBN, as shown in Figure 3. For each slice, we create two "mega nodes" that contains all the (hidden) nodes in that slice. The messages coming into the first mega node are assumed independent; they are then multiplied together to form the (approximate) prior[3]; a single message is then sent to the second mega node, corresponding to an exact update step using the $Q^N \times Q^N$ transition matrix; finally, the individual marginals are computed, and the process is repeated. Of course, BK does not actually construct the mega nodes, and does the exact update using junction tree, but the two algorithms are functionally equivalent. To simulate BK when the clusters contain more than one node, we simply create new clustered nodes, in addition to the mega-nodes, and run LBP on the new graph, as illustrated in Figure 3.

Since FF and BK are equivalent to one iteration of LBP, on the regular and clustered graphs respectively, we can improve on both of these algorithms by iterating more than once. This gives the algorithm the opportunity to "recover" from its incorrect independence assumptions. We will see in the Section 5 that even a small number of iterations can help dramatically.

[2]In the case of noisy-or nodes, there are efficient ways to compute the $\lambda$ and $\pi$ messages without having to do work which is exponential in the number of parents [16]. This reduces the overall complexity of FF from $O(TNQ^{F+1})$ to $O(TNFQ)$.

[3]For a directed graph, naive Pearl would take $O(Q^N)$ time to compute $\pi$ for the mega-node, but we can do this in $O(QN)$ time by exploiting the fact that the CPT factorizes. Alternatively, we can use an undirected graph, in which the computation of messages always takes time linear in the number of neighbors.

### 4.1 A free energy for iterated BK

The equivalence between BK and a single iteration of LBP on the clustered graph allows us to utilize the recent result of Yedidia et al [22] to obtain a free energy for "iterated" BK. We define the "iterated" BK algorithm as running LBP on the clustered graph using a FB schedule until convergence. The first iteration of iterated BK is equivalent to BK but in subsequent iterations, the $\alpha$ and $\beta$ messages interact to improve the quality of approximation. The analysis of [22] shows that iterated BK can only converge to zero gradient points of the Bethe free energy.

This sheds light over the relationship between iterated BK and the mean field (MF) approximation. The MF free energy is the same as the iterated BK free energy when joint distributions over pairs of nodes are replaced by a product of marginal beliefs over individual nodes: iterated BK captures dependencies between nodes in subsequent slices while MF does not. While this result only holds for iterated BK, ordinary BK can be thought of as a first approximation to iterated BK.

## 5 Experimental results

In this section, we compare the BK algorithm with $k$ iterations of LBP on the original graph, using the FB protocol ($k = 1$ iteration corresponds to FF). We used a CHMM model with 10 chains trained on some real freeway traffic data using exact EM [12]. We define the $L_1$ error as $\Delta_t = \sum_{i=1}^{N} \sum_{s=1}^{Q} |P(X_t^i = s|y_{1:T}) - \hat{P}(X_t^i = s|y_{1:T})|$, where $P(\cdot)$ is the exact posterior and $\hat{P}(\cdot)$ is the approximate posterior. In Figure 4, we plot this against $t$ for 1–4 iterations of LBP. Clearly, the posteriors are oscillating, and this happens on many sequences with this model. We therefore used the damping trick described in [15]. In this case, each new message is defined to be a convex combination of the usual expression and the old messsage, with weight $\mu$ given to the old message. Hence $\mu = 0$ corresponds to undamped propagation, and $\mu = 1$ corresponds to not updating the messages at all, i.e., only using local evidence. It is easy to show that any fixed points reached using this algorithm are fixed points of the original set of (undamped) equations. It is clear from Figure 5 that damping helps considerably. The results are summarised in Figure 6, where we see that after a small number of iterations, LBP with $\mu = 0.1$ is doing better than BK. Other sequences give similar behavior.

To check that these results are not specific to this model/ data set, we also compared the algorithms on the water DBN shown in Figure 1. We generated observation sequences of length 100 from this model us-ing random parameters and binary nodes, and then compared the marginal posteriors as a function of number of iterations and damping factor. The results for a typical sequence are shown in Figure 7. This time we see that there is no oscillation, and that as few as two iterations of LBP can outperform BK.

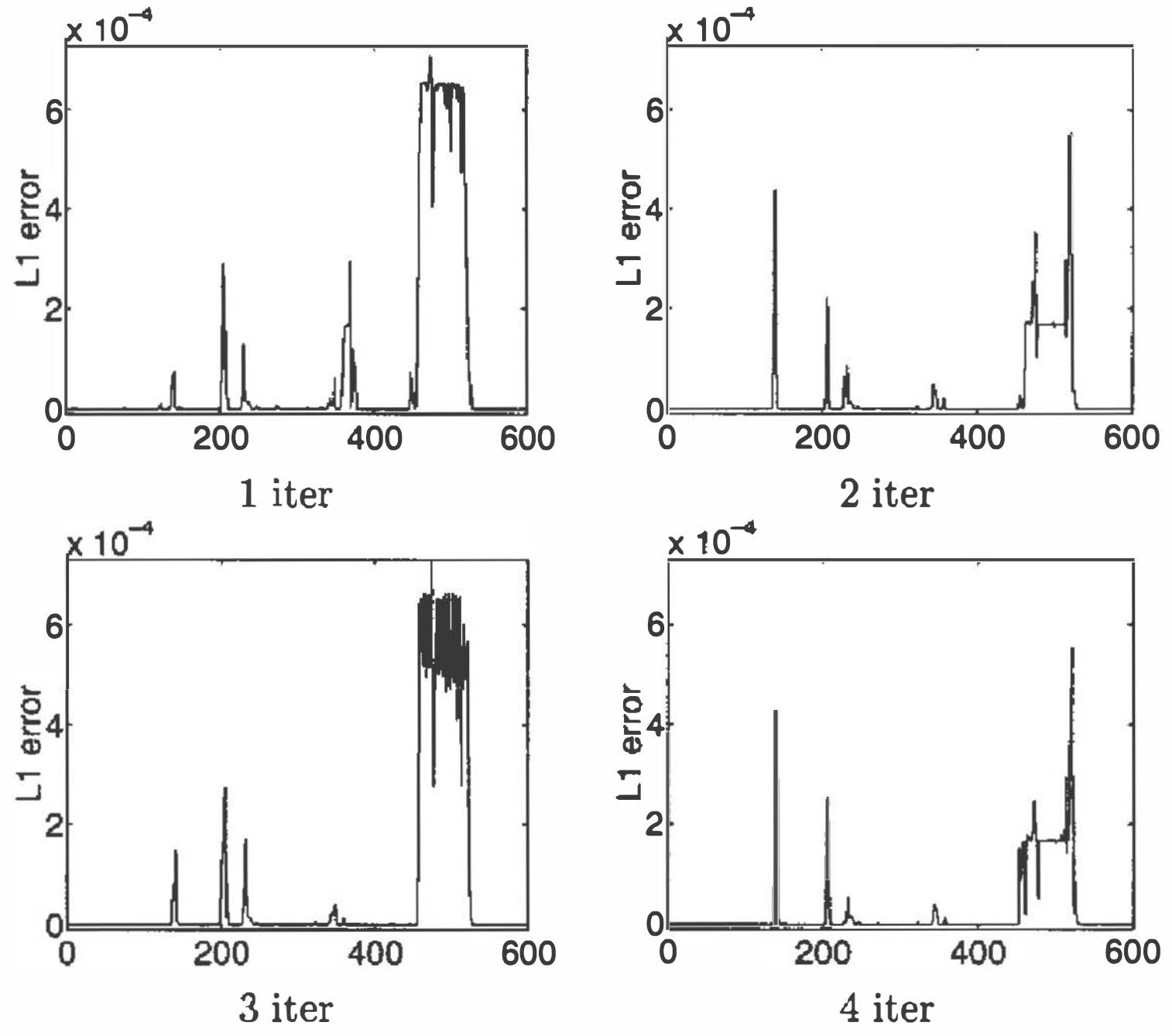

Figure 4: $L_1$ error on marginal posteriors vs. timeslice after iterations 1–4 of undamped LBP applied to the traffic CHMM. The $L_1$ error oscillates with a period of 2 (as seen by the similarity between the graphs for iterations 1/3 and 2/4); this implies that the underlying marginals are oscillating with the same period.

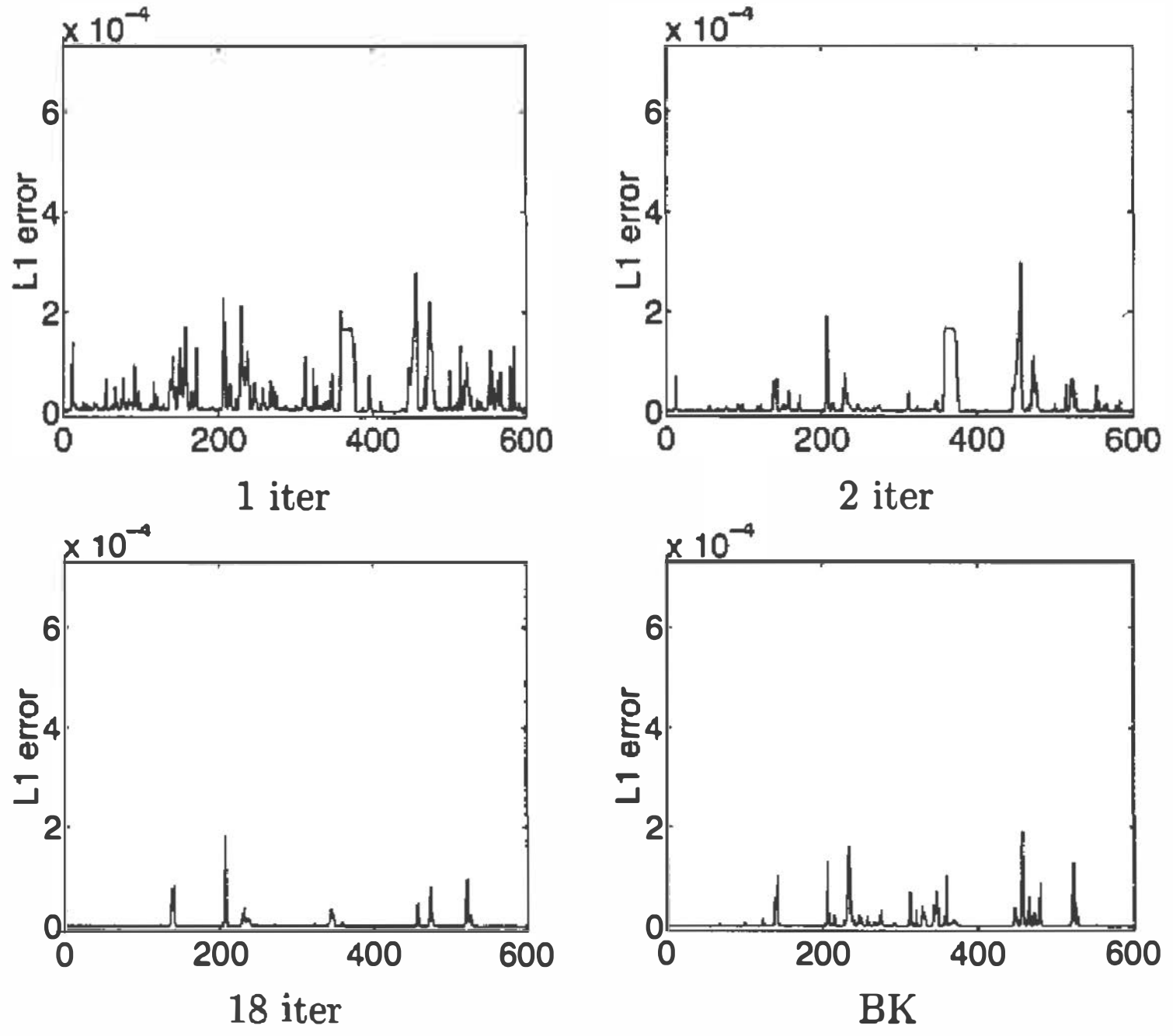

Figure 5: iterations 1, 2, and 18 of LBP with damping factor $\mu = 0.1$, and after using 1 iteration of BK, on the traffic CHMM.

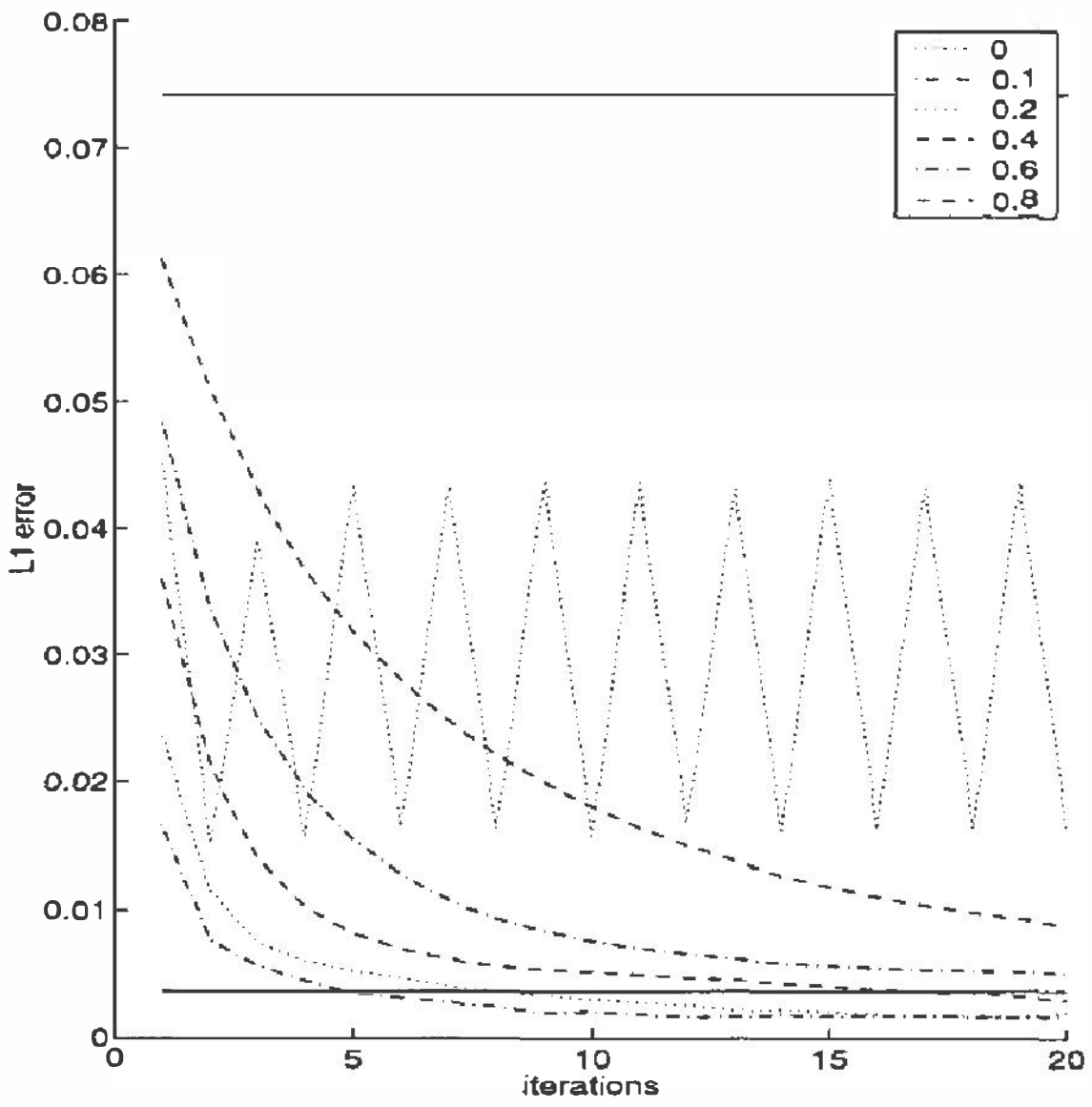

Figure 6: Results of applying LBP to the traffic CHMM with 10 chains. The lower solid horizontal line is the error incurred by BK. The oscillating line is the error incurred by LBP using damping factor $\mu = 0$; the lowest curve corresponds to $\mu = 0.1$ and the highest to $\mu = 0.8$. The upper horizontal line corresponds to not updating the messages at all ($\mu = 1$), and gives an indication of performance based on local evidence alone.

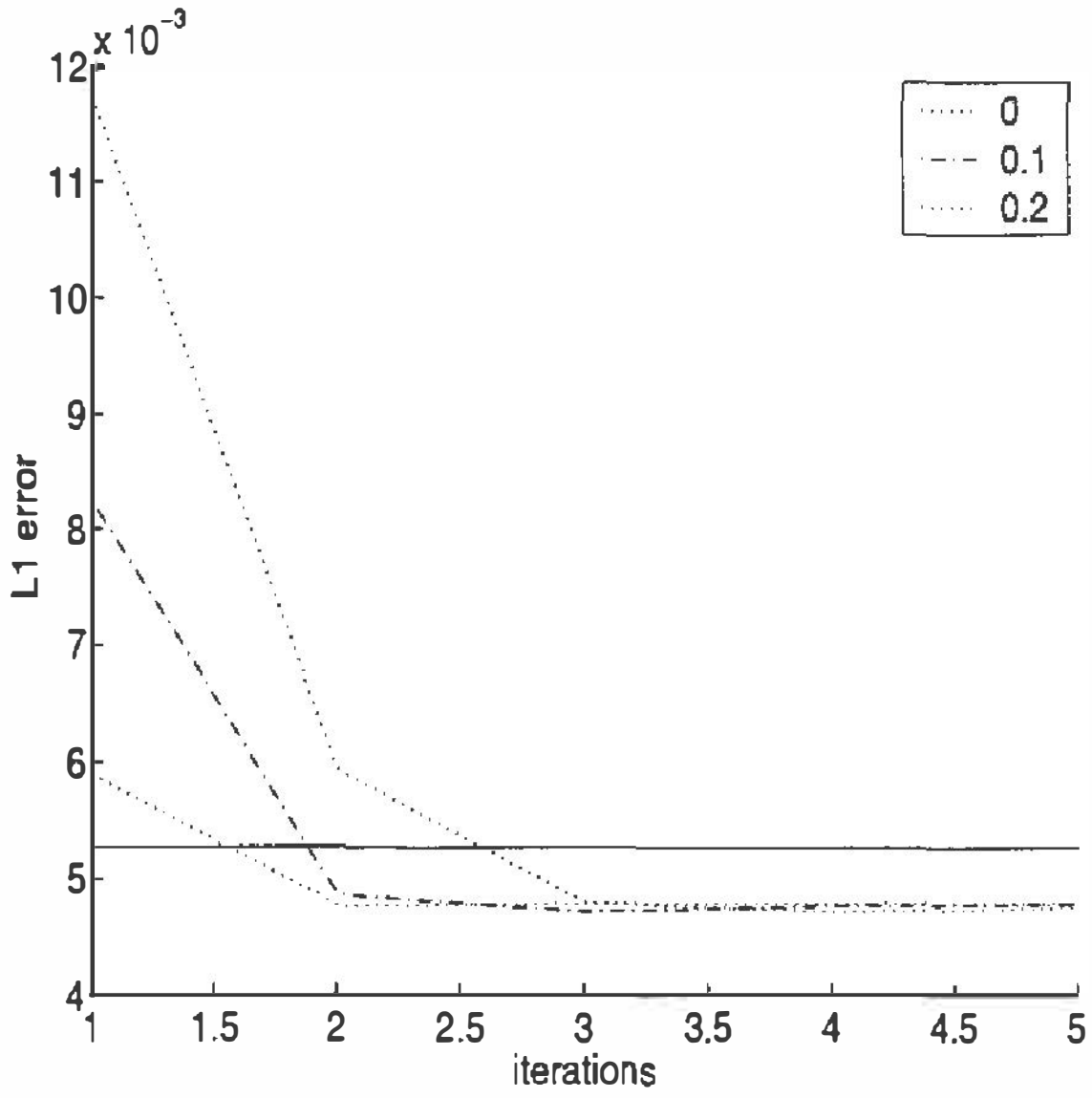

Figure 7: Same as Figure 6, but for the water network. The dotted lines, from top to bottom, represent $\mu = 0$, $\mu = 0.1$ and $\mu = 0.2$. The solid line represents BK.

In addition to accuracy, it is also interesting to see how the algorithms compare in terms of speed. We therefore generated random data from CHMMs with $N = 1, 3, \ldots, 11$ chains, and computed the posteriors using the different methods. The running times are shown in Figure 8. It is clear that both BK and FF/LBP have a running time which is linear in $N$ (for the CHMM model), but the constant factors of BK are much higher, due to the complexity of the algorithm, and in particular, the need to perform repeated marginalisations. This is also why BK is slower than exact inference for $N < 11$, even though it is asymptotically more efficient.[4]

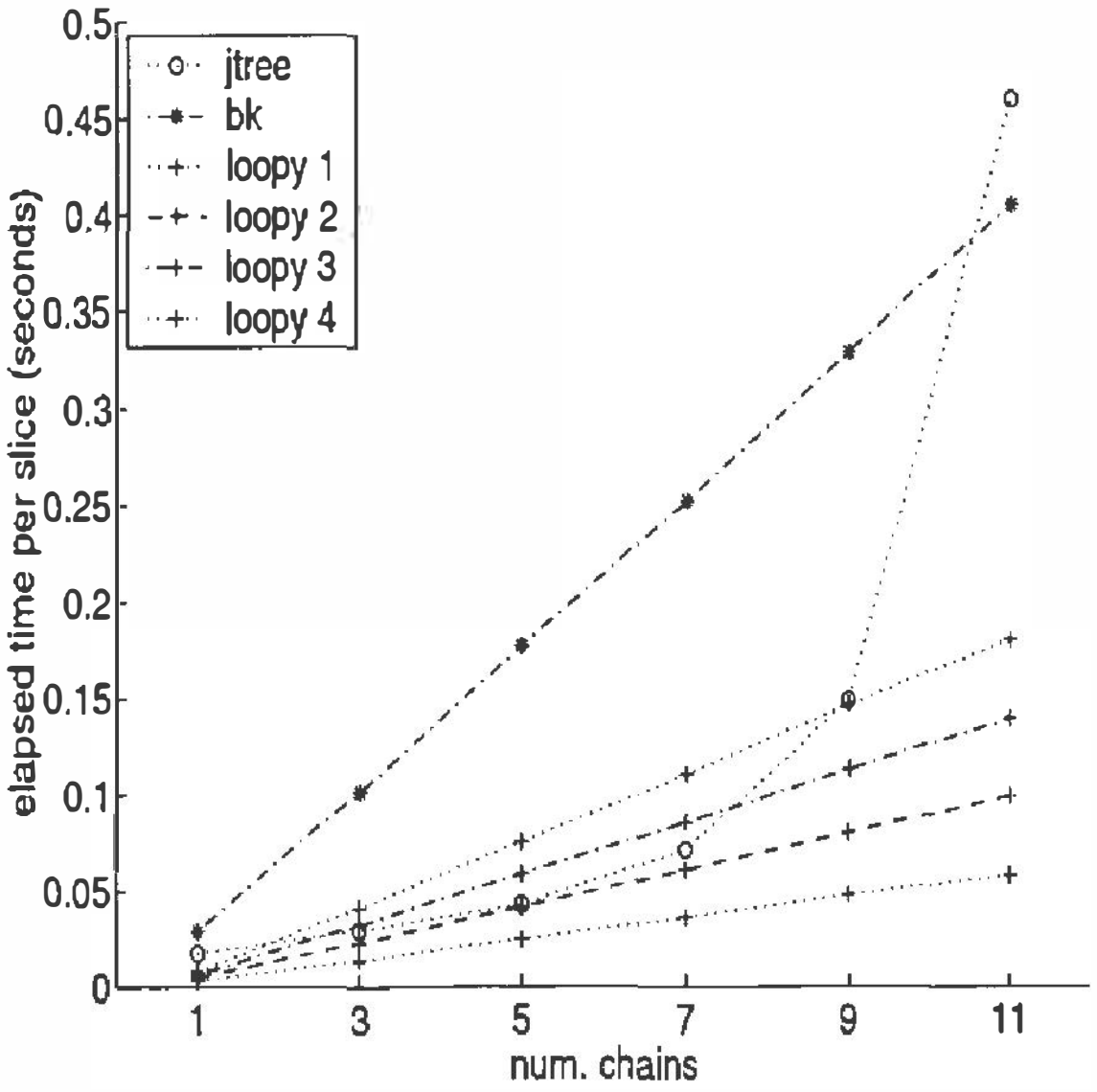

Figure 8: Running time on CHMMs as a function of the number of chains. The vertical axis is the total running time divided by the length of the sequence. The horizontal axis is the number of chains. The dotted curve is junction tree, the steep straight line is BK, and the shallow straight lines are LBP; "loopy $k$" means $k$ iterations of LBP.

[4]All algorithms were implemented in Matlab and are included in the Bayes Net Toolbox, which can be downloaded from www.cs.berkeley.edu/ ~murphyk/Bayes/bnt.html.

## 6 Related work

We have already discussed in detail the connections between LBP, BK and FF. However, there are several other approximate inference algorithms with a very similar "flavor". Perhaps the closest is the expectation propagation algorithm [14]. This is also an iterative message passing algorithm, but now the messages encode moments of the variables computed with respect to some approximating distribution. The mini-bucket algorithm [4] also approximates joint distributions over collections of nodes as a product of smaller terms; however, this is not an iterative algorithm, and hence cannot correct for erroneous independence assumptions made in the first pass.

## 7 Conclusions

We have described a very simple approximate inference algorithm for DBNs, and shown that it is equivalent to a single iteration of loopy belief propagation (LBP). We have also elucidated the connection between the BK algorithm and LBP, and we used the free energy of LBP to compare BK and the mean field approximations. Finally, we showed empirically that LBP can improve the accuracy of both FF and BK.

### Acknowledgments

We would like to thank the UAI reviewers for their helpful comments, and the following grants for financial support: MURI ARO-DAAH04-96-1-0341, MURI N00014-00-1-0637 and NSF IIS-9988642.

## References

[1] X. Boyen and D. Koller. Approximate learning of dynamic models. In *NIPS-11*, 1998.

[2] X. Boyen and D. Koller. Tractable inference for complex stochastic processes. In *UAI*, 1998.

[3] R. G. Cowell, A. P. Dawid, S. L. Lauritzen, and D. J. Spiegelhalter. *Probabilistic Networks and Expert Systems*. Springer, 1999.

[4] R. Dechter. Mini-buckets: a general scheme of approximating approximations in automated reasoning. In *IJCAI*, 1997.

[5] W. Freeman and Y. Weiss. On the fixed points of the max-product algorithm. *IEEE Trans. on Info. Theory*, 2000. To appear.

[6] B. J. Frey. Turbo factor analysis. *Neural Computation*, 1999. Submitted.

[7] Z. Ghahramani and M. Jordan. Factorial hidden Markov models. *Machine Learning*, 29:245–273, 1997.

[8] F. V. Jensen, U. Kjaerulff, K. G. Olesen, and J. Pedersen. An expert system for control of waste water treatment — a pilot project. Technical report, Univ. Aalborg, Judex Datasystemer, 1989. In Danish.

[9] U. Kjaerulff. Triangulation of graphs - algorithms giving small total state space. Technical Report R-90-09, Dept. of Math. and Comp. Sci., Aalborg Univ., Denmark, 1990.

[10] U. Kjaerulff. A computational scheme for reasoning in dynamic probabilistic networks. In *UAI-8*, 1992.

[11] J. Kwon and K. Murphy. Modeling freeway traffic with coupled HMMs. Technical report, Univ. California, Berkeley, 2000.

[12] R. J. McEliece, D. J. C. MacKay, and J. F. Cheng. Turbo decoding as an instance of Pearl's 'belief propagation' algorithm. *IEEE J. on Selectred Areas in Comm.*, 16(2):140–152, 1998.

[13] T. Minka. A family of algorithms for approximate Bayesian inference. In *UAI*, 2001.

[14] K. Murphy, Y. Weiss, and M. Jordan. Loopy belief propagation for approximate inference: an empirical study. In *UAI*, 1999.

[15] J. Pearl. *Probabilistic Reasoning in Intelligent Systems: Networks of Plausible Inference*. Morgan Kaufmann, 1988.

[16] M. Peot and R. Shachter. Fusion and propogation with multiple observations in belief networks. *Artificial Intelligence*, 48:299–318, 1991.

[17] L. R. Rabiner. A tutorial on Hidden Markov Models and selected applications in speech recognition. *Proc. of the IEEE*, 77(2):257–286, 1989.

[18] P. Smyth, D. Heckerman, and M. I. Jordan. Probabilistic independence networks for hidden Markov probability models. *Neural Computation*, 9(2):227–269, 1997.

[19] Y. Weiss. Correctness of local probability propagation in graphical models with loops. *Neural Computation*, 12:1–41, 2000.

[20] Y. Weiss and W. T. Freeman. Correctness of belief propagation in Gaussian graphical models of arbitrary topology. In *NIPS-12*, 1999.

[21] J. Yedidia, W. T. Freeman, and Y. Weiss. Generalized belief propagation. In *NIPS-13*, 2000.

[22] G. Zweig. A forward-backward algorithm for inference in Bayesian networks and an empirical comparison with HMMs. Master's thesis, Dept. Comp. Sci., U.C. Berkeley, 1996.